\documentclass{llncs}

\usepackage{graphicx}
\usepackage[colorinlistoftodos]{todonotes}
\usepackage{algorithm}
\usepackage{algorithmic}
\usepackage{mathtools}
\usepackage{amsmath}
\usepackage{array}

\begin{document}

\title{Prioritized Sweeping Neural DynaQ with Multiple Predecessors, and Hippocampal Replays}
\titlerunning{DynaQ \& Replays}  
%
\author{Lise Aubin \and Mehdi Khamassi \and Beno{\^i}t Girard}
\authorrunning{Aubin, Khamassi \& Girard} 
\institute{Sorbonne Universit\'{e}, CNRS, Institut des Syst{\`e}mes Intelligents \\ et de Robotique (ISIR), F-75005 Paris, France\\
\email{benoit.girard@sorbonne-universite.fr}
}

\maketitle              

\begin{abstract}
During sleep and wakeful rest, the hippocampus replays sequences of place cells that have been activated during prior experiences. 
These replays have been interpreted as a memory consolidation process, but recent results suggest a possible interpretation in terms of reinforcement learning.
The \emph{Dyna} reinforcement learning algorithms use off-line replays to improve learning. Under limited replay budget, \emph{prioritized sweeping}, which requires a model of the transitions to the predecessors, can be used to improve performance. We investigate if such algorithms can explain the experimentally observed replays.
We propose a neural network version of prioritized sweeping Q-learning, for which we developed a growing multiple expert algorithm, able to cope with multiple predecessors. 
The resulting architecture is able to improve the learning of simulated agents confronted to a navigation task. We predict that, in animals, learning the transition and reward models should occur during rest periods, and that the corresponding replays should be shuffled.

\end{abstract}
\section{Introduction}
The hippocampus hosts a population of cells responsive to the current position of the animal within the environment, the place cells (PCs), a key component of the brain navigation system \cite{Okeefe71}. Since the seminal work of Wilson et al. \cite{Wilson1994}, it has been shown that PCs are reactivated during sleep -- obviously in the absence of locomotor activity -- and that these reactivations are functionally linked with improvement of the learning performance of a navigation task \cite{girardeau2009}. Similar reactivations have been observed in the a wakeful state \cite{Foster2006}, while the animal is immobile, either consuming food at a reward site, waiting at the departure site for the beginning of the next trial or stopped at a decision point. These reactivations contain sequences of PCs' activations experienced in the wakeful state (forward reactivations) \cite{Lee2002}, sequences played in the reverse order (backward reactivations) \cite{Foster2006}, and sometimes novel sequences (resulting from the concatenation of experienced sequences) \cite{Gupta2010}. 
These reactivations have been interpreted in the light of the memory consolidation theory \cite{chen2017}: they would have the role of copying volatile hippocampal memories into the cortex \cite{peyrache2009} for reorganization and longer-term storage \cite{mcclelland1995}. However, recent results have shown that these reactivations also have a causal effect on reinforcement learning processes \cite{girardeau2009,delavilleon2015}.

\begin{figure}[t]
\centering
\includegraphics[width=.55\textwidth]{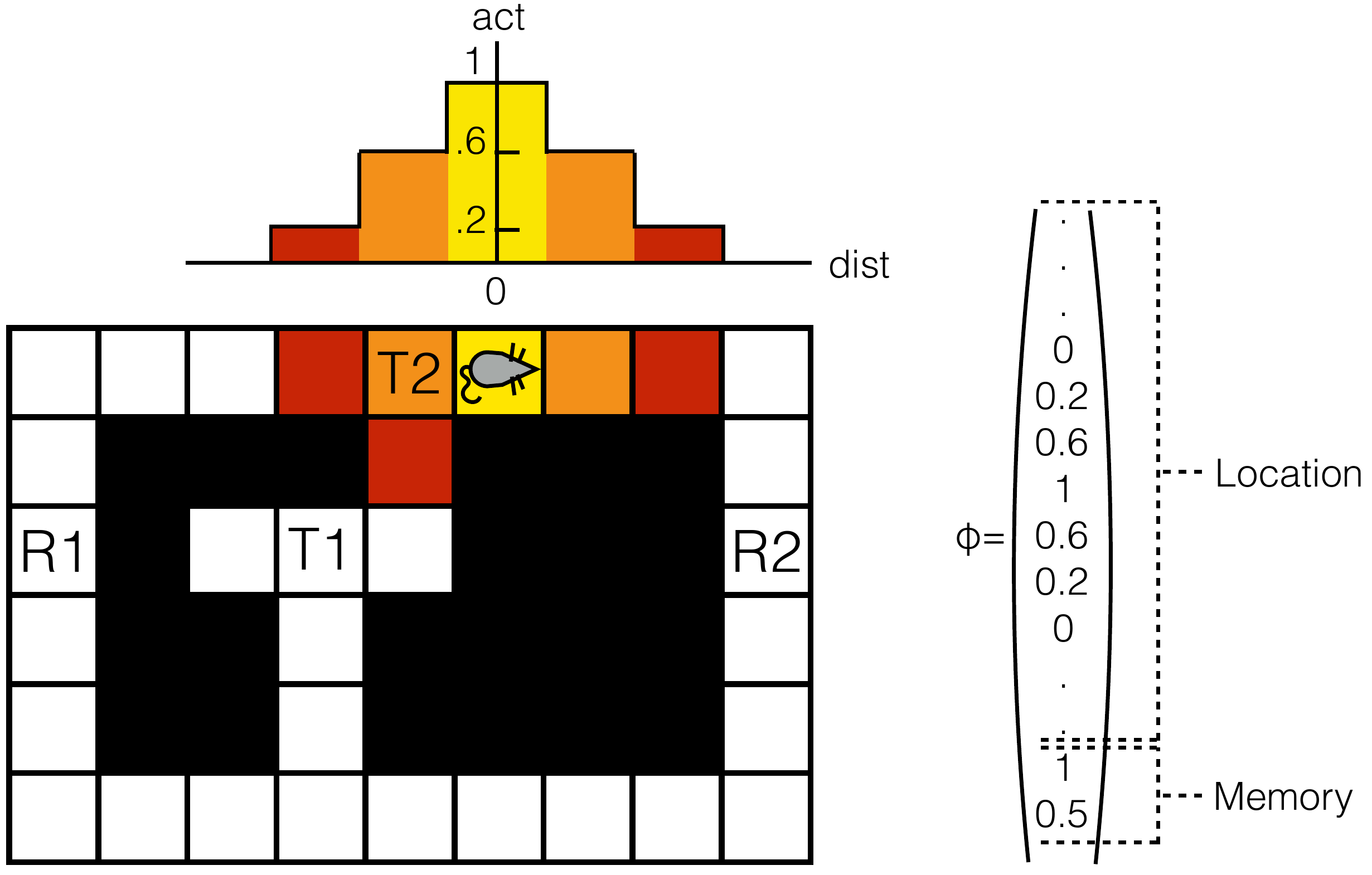}
\caption{\label{fig:task}
\textbf{Model of the rat experiment used in \cite{Gupta2010}.} The maze is discretized into 32 positions (squares). The agent can use 4 discrete actions (N,E,S,W). The input state $\phi$ is the concatenation of 32 location components and two reward memory components. 
The location part of $\phi$ represents the activation of 32 place cells co-located with the maze discrete positions, their activity $act$ depends on the Manhattan distance of the agent to the cell. Figures 1 to 5 by Aubin \& Girard, 2018; available at https://doi.org/10.6084/m9.figshare.5822112.v2 under a CC-BY4.0 license.
%
%
}
\end{figure}

A number of reinforcement learning (RL) algorithms make use of input reactivations, reminiscent of hippocampal reactivations. These algorithms are thus candidates to explain the replay phenomenon \cite{Caze18}. Among them, the Dyna family of algorithms \cite{sutton1990} is of special interest because it was specifically designed to make the best possible use of alternation between on-line and off-line learning phases (i.e. phases during which the agent acts in the real world or in simulation). We concentrate here on the Q-learning version of Dyna (Dyna-Q). When operating on-line, Dyna-Q is indistinguishable from the original \emph{model-free} Q-learning algorithm: it computes reward prediction error signals, and uses them to update the estimated values of the (state, action) couples, $Q(s,a)$. 
In its original version \cite{sutton1990}, when off-line, the Dyna algorithm reactivates randomly chosen quadruplets composed of an initial state, a chosen action, and the predicted resulting state and reward (produced by a learned world-model, this phase being thus \textit{model-based}), in order to refine the on-line estimated values. However, when the number of reactivations is under a strict budget constraint, it is more efficient to select those that will provide more information, which are those that effectively generated a large reward prediction error in the last on-line phase, and those that are predicted to do so by the world model, a principle called \emph{prioritized sweeping} \cite{moore1993,peng1993}.

We are interested here in mimicking the process by which the basal ganglia, which is central for RL processes \cite{khamassi2005}, can use the state representations of the world that are provided by the hippocampus. The manipulated state descriptor will thus be a population activity vector, 
and we will represent the Q-values and the world model with neural network approximators \cite{lin1992}.

In the following, we describe the rat experimental setup used in \cite{Gupta2010}, and how we simulated it. In this task, a state can have multiple predecessor states resulting from a single action
, we thus present a modified Dyna-Q learning algorithm, with a special emphasis on the neural-network algorithm we designed to learn to approximate binary relations (not restricted to functions) with a \emph{growing} approach: GALMO for Growing Algorithm to Learn Multiple Outputs. Our results successively illustrate three main points. First, because of interferences between consecutively observed states during maze experience, themselves due to the use of a neural-network function approximator, the world model had to be learned with shuffled states during off-line replay. Second, GALMO allows to efficiently solve the multiple predecessor problem. Third, the resulting system, when faced with a training schedule similar to \cite{Gupta2010}, generates a lot of disordered state replays, but also a non-negligible set of varied backward and forward replay sequences, without explicitly storing and replaying sequences.

\section{Methods}
\subsection{Experimental task}

\paragraph{}We aim at modeling the navigation task used in \cite{Gupta2010}: two successive T-mazes (T1 and T2 on \textbf{Fig.~\ref{fig:task}}), with lateral return corridors. The left and right rewarding sites deliver food pellets with different flavors. The training involves daily changing contingencies, forcing rats to adapt their choice to turn either left or right at the final choice (T2) based on the recent history of reward. These contingencies are: 1) always turn right, while the left side of the maze is blocked; 2) always turn left, while the right side of the maze is blocked; 3) always turn right; 4) always turn left; 5) alternate between left and right on a lap-by-lap basis.

Rats attempting to run backward in the maze were prevented from doing so by the experimenter. The first day, they were exposed to task 1 (40 trials), and the next day, to task 2 (40 trials also). Then, depending on their individual learning speed, rats had between seventeen and twenty days to learn tasks 3, 4 and 5 (a single condition being presented each day). Once they had reached at least 80\% success rate on all tasks, rats were implanted with electrodes; after recovery, recording sessions during task performance lasted for six days.

During the six recording sessions, the reward contingency was changed approximately midway through the session and hippocampal replays were analyzed when rats paused at reward locations.
Original analyses of replayed sequences \cite{Gupta2010} revealed that: during same-side replays (i.e., replays representing sequences of previously visited locations on the same arm of the maze as the current rat position) forward and backward replays started from the current position; during opposite-side replays (i.e., representing locations on the opposite arm of the maze) forward replays occurred mainly on the segment leading up to reward sites, and backward replays covered trajectories ending near reward sites. In general, the replay content did not seem to only reflect recently experienced trajectories, since trajectories experienced 10 to 15 minutes before were replayed as well. Indeed, there were more opposite-side replays during task 3 and 4 than during the alternation task. 
Finally, among all replays, a few were shortcuts never experienced before which crossed a straight path on the top or bottom of the maze between the reward sites.

\subsection{Simulation}

\paragraph{}We have reproduced the T-maze configuration with a discrete environment composed of 32 squares (\textbf{Fig.~\ref{fig:task}, left}), each of them representing a $10\times10$ cm area. States are represented by a vector $\phi$, concatenating place cells activity and a memory of past rewards (\textbf{Fig.~\ref{fig:task}, right}). The modeled place cells are centered on the discrete positions and their activity (color-coded on \textbf{Fig.~\ref{fig:task}}) decreases with the Manhattan distance between the simulated rat position to the position they encode (top of \textbf{Fig.~\ref{fig:task}}). When a path is blocked (contingencies 1 and 2), the activity field does not expand beyond walls and will thus shrink, as is the case for real place cells \cite{paz2004}. 
To represent the temporal dimension\textcolor{blue}{,} which is essential during the alternation task, we have added two more components in the state's vector representation (\textbf{Fig.~\ref{fig:task}, right}): the left side reward memory (L) and the right side reward memory (R). They take a value of $1$ if the last reward was obtained on that side, $0.5$ if the penultimate reward was on that side, and $0$ if that side has not been rewarded during the last two reward events. Therefore, after two successful laps, the task at hand can be identified by the agent based on the value of this memory (\textbf{Tab.~\ref{tab:mem}}).
This ability to remember the side of the rewards is supposed to be anchored both on the different position and flavor cues that characterize each side.
Since it has been shown that, beyond purely spatial information, the hippocampus contains contextual information important for the task at hand \cite{eichenbaum2017}, we hypothesize that this memory is also encoded within the hippocampus, along with the estimation of the agent's current position.

The agent can choose between four actions: North, South, East and West. As in the real experiment, the agent cannot run backward.
\begin{table}
\caption{State of the L and R memory components of $\phi$ and corresponding meaning in terms of task at hand, after two successful laps.\label{tab:mem}}
\centering
\begin{tabular}{ccl}
  \hline
  \textbf{L} & \textbf{R} & \textbf{Task identification (after 2 laps)} \\
  \hline
  1 & 0 & Always turn right (Tasks 1 \& 3)\\
  0 & 1 & Always turn left (Tasks 2 \& 4) \\
	0.5 & 1 & Alternation (Task 5), go left next time \\
	1 & 0.5 & Alternation (Task 5), go right next time\\
  \hline
\end{tabular}
\end{table}
\subsection{Neural DynaQ with a prioritized sweeping algorithm}

\paragraph{}Our algorithm is based on a Dyna architecture \cite{sutton1990} which means that, as in model-based architectures, a world model composed of a reward and a transition model has to be learned \cite{Sutton1998}. \emph{Prioritized sweeping} \cite{moore1993,peng1993} requires the transition model to allow the prediction of the predecessors of a state $s$ given an action $a$, because this information will be needed to back-propagate the reward prediction computed in state $s$ to its predecessors. Hence, our architecture is composed of two distinct parts: one dedicated to learning the world model, and the other one to learning the Q-values.

%
\begin{algorithm}
\caption{LearnWM: learn the world model}
\label{algo:LearnWM}
\begin{algorithmic}
  \STATE collect $\mathcal{S}$ // a set of $(\phi^t, \phi^{t-1}, a, r)$ quadruplets
  \FOR{$k \in \{N,S,E,W\}$}
    \STATE $\mathcal{S}_P^k \leftarrow \{(\phi^t,\phi^{t-1}):(\phi^t, \phi^{t-1}, a, r) \in \mathcal{S} \text{ and } a=k\}$
    \STATE $\mathcal{S}_R^k \leftarrow \{(\phi^t,r):(\phi^t, \phi^{t-1}, a, r) \in \mathcal{S} \text{ and } a=k\}$
    \FOR{$f \in \{P,R\}$}
    \STATE // P,R: Predecessor and Reward types of networks
      \STATE $\mathcal{N}_f^{k} \leftarrow$ null // list of networks (outputs)
      \STATE $\mathcal{G}_f^{k} \leftarrow$ null // list of networks (gates)
      \STATE create $N_{new}^{k}$ ; append $N_{new}^{k}$ to $\mathcal{N}_f^{k}$
      \STATE create $G_{new}^{k}$ ; append $G_{new}^{k}$ to $\mathcal{G}_f^{k}$
      \STATE GALMO($\mathcal{S}_f^k$, $\mathcal{N}_f^{k}$, $\mathcal{G}_f^{k}$) // refer to Algo~\ref{algo:galmo} for this specific training procedure
    \ENDFOR
  \ENDFOR
\end{algorithmic}
\end{algorithm}

\textbf{Learning the world model.} Two sets of neural networks compose the world model. 
Four reward networks $N_{R}^{a}$, one for each action $a$, learn the association between (state, action) couples and rewards ($N_{R}^{a}: s \rightarrow r(s,a)$). Four other networks $N_{P}^{a}$ learn the states for which a transition to a given state $s$ is produced after execution of action $a$, i.e., the predecessors of $s$ ($N_{P}^{a}:s\rightarrow \{s'\}$). 

Owing to the nature of the task (navigation constrained by corridors) and the states' representation, the data that must be learned are not independent. Indeed, successive state vectors are very similar due to the overlap between place-fields, and are always encountered in the same order during tasks execution (because the agent always performs the same stereotyped trajectories along the different corridors). However, it is well known that the training of a neural network is guaranteed to converge only if there is no correlation in the sequence of samples submitted during learning, a condition that is often not respected when performing on-line reinforcement learning \cite{tsitsiklis1997}. We indeed observed that in the task at hand, despite its simplicity, it was necessary to store the successive observations and to train the world model off-line with a shuffled presentation of the training samples (for the general scheme of the off-line training, see \textbf{Algo.~\ref{algo:LearnWM}}). For that reason, we created a dataset $\mathcal{S}$ compiling all transitions, i.e ($\phi^t$, $\phi^{t-1}$, a, r) quadruplets from all tasks. When there is no predecessor of $\phi^t$ by action $a$ (as can be the case when this action would require to come through a wall), the transition is represented as ($\phi^t$, $\vec{0}$, a, r): those "null" transitions allow $N_{P}^{a}$ networks to represent the fact that the transition does not exist.

\begin{figure}
\centering
\includegraphics[width=.8\textwidth]{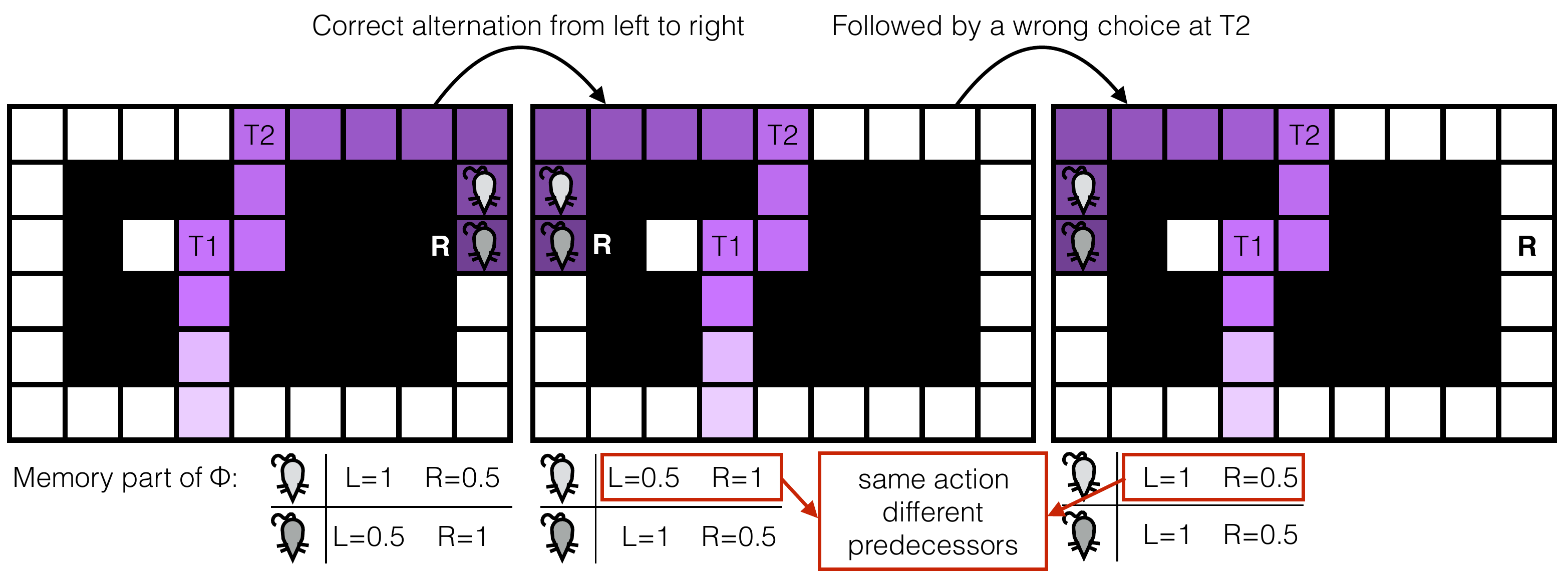}
\caption{\label{fig:multpred}
\textbf{Example of multiple predecessors in the alternation task.} The agent first correctly goes to the right (left). It then goes to the left (middle) where, at the reward site, its predecessor state has a $(L=0.5,R=1)$ memory component. It then makes a wrong decision and goes to the left again (right), but is not rewarded: at this last position, the location component of the predecessor state (white mouse) is identical but the memory component is different $(L=1,R=0.5)$ from the previous lap. Violet gradient: past trajectory; white mouse: previous position; gray mouse: current position; white R: agent rewarded; black R: current position of the reward.}
\end{figure}

Despite its simplicity, the navigation task modeled here has some specificities: during task 5 (alternation), some states have more than one predecessor for a given action (see an example on \textbf{Fig.~\ref{fig:multpred}}), the algorithm must thus be capable of producing more than one output for the same input. To do that, we have created a growing type of algorithm inspired by \textit{mixture of expert} algorithms \cite{jacobs1991} (which we call here the \textit{GALMO} algorithm, see \textbf{Algo.~\ref{algo:galmo}}), based on the following principles:
\begin{itemize}
  \item The algorithm should allow the creation of multiple $N_i$ networks (if needed) so that a single input can generate multiple outputs. Each of these networks is coupled with a gating network $G_i$, used after training to know if the output of $N_i$ has to be taken into account when a given sample is presented.
  \item When a sample is presented, the algorithm should only train the $N_i$ network that generates the minimal error (to enforce network specialization), and remember this training event by training $G_i$ to produce $1$ and the other $G_{k\neq i}$ to produce $0$.
  \item The algorithm should track the statistics of the minimal training errors of each sample during an epoch, so as to detect samples whose error is much higher than the others'. GALMO assumes that these outliers are caused by inputs which should predict multiple outputs and which are stuck in predicting the barycenter of the expected outputs. A sample is considered an outlier when its error is larger than a threshold $\theta$, equal to the median of the current error distribution, plus $w$ times the amplitude of the third quartile ($Q3-median$). When such a detection occurs, a new network is created on the fly, based on a copy of the network that produced the minimal error for the sample. The new network is then trained once on the sample at hand.
\end{itemize}

\begin{algorithm}
\caption{GALMO: Growing algorithm to learn multiple outputs}
\label{algo:galmo}
\begin{algorithmic} 

\STATE \textbf{INPUT:} $\mathcal{S}$, $\mathcal{N}$, $\mathcal{G}$
\STATE \textbf{OUTPUT:} $\mathcal{N}$, $\mathcal{G}$
\STATE // $\mathcal{S} = \langle (in_0,out_0), ..., (in_n,out_n) \rangle$ : list of samples
\STATE // $\mathcal{N} = \langle N_0 \rangle$ : lists of neural networks (outputs)
\STATE // $\mathcal{G} = \langle G_0 \rangle$ : lists of neural networks (gates)
\STATE $\theta \leftarrow +\infty$\\

\FOR{nbepoch $\in \{ 1, maxepoch\}$}
  \STATE $\mathcal{M}$ $\leftarrow$ null // $\mathcal{M}$ is a list of the minimal error per sample
  \FOR{each (in,out)$\in$ $\mathcal{S}$}
    \STATE $\mathcal{E}$ $\leftarrow$ null // $\mathcal{E}$ is a list of errors for a sample
    \FOR{each $N\in$ $\mathcal{N}$}
  	  \STATE append $\lVert N(in)-out \rVert_{L_{1}}$ to $\mathcal{E}$
    \ENDFOR
	\IF{$\min(\mathcal{E}) < \theta$} 
	  \STATE backprop($N_{argmin(\mathcal{E})}, in, out$) \\
	  \STATE backprop($G_{argmin(\mathcal{E})}, in, 1$) \\
	  \FOR {each $G \in \mathcal{G}$ with $G \ne G_{argmin(\mathcal{E})}$} 
		\STATE backprop($G, in, 0$)
      \ENDFOR
	\ELSE
	  \STATE create $N_{new}$; append $N_{new}$ to $\mathcal{N}$
      \STATE $N_{new} \leftarrow$ copy$(N_{argmin(\mathcal{E})})$
	  \STATE backprop($N_{new}$, $input=$in, $target=$out) 
	  \STATE create $G_{new}$; append $G_{new}$ to $\mathcal{G}$
	  \STATE backprop($G_{new}, in, 1$) 
    \ENDIF
  \ENDFOR
  \STATE $\theta \leftarrow median(\mathcal{M}) + w*(Q3(\mathcal{M}) - median(\mathcal{M}))$
\ENDFOR
\end{algorithmic}
\end{algorithm}

In principle, the algorithm could be modified to limit the maximal number of created networks, or to remove the networks that are not used anymore, but these additions were not necessary here.
%
%
\begin{algorithm}[t]
\caption{Neural Dyna-Q with \textit{prioritized sweeping} \& multiple predecessors}
\label{algo:dyna}
\begin{algorithmic} 

\STATE \textbf{INPUT:} $\phi^{t=0}$, $\mathcal{N_P}$, $\mathcal{G_P}$, $\mathcal{N_R}$, $\mathcal{G_R}$
\STATE \textbf{OUTPUT:} $N_Q^{a \in \{ N,S,E,W\}}$

\STATE PQueue $\leftarrow \{ \}$ // PQueue: empty priority queue
\STATE nbTrials $\leftarrow 0$
\REPEAT
  \STATE a $\leftarrow$ softmax($N_{Q}(\phi^{t})$) 
  \STATE take action a, receive r, $\phi^{t+1}$ 
  \STATE backprop($N_{Q}^{a}$, $input = \phi^{t}$, $target = r + \gamma max_a(N_{Q}(\phi^{t+1})$) 

  \STATE Put $\phi^{t}$ in PQueue with priority $|N_{Q}^{a}(\phi^{t}) - (r + \gamma max_a(N_{Q}(\phi^{t+1})))|$ 

    \IF{r$>0$}
      \STATE nbReplays $\leftarrow 0$
      \STATE Pr $= \langle \rangle$ // empty list of predecessors
      \REPEAT 
 	    \STATE $\phi \leftarrow$ pop(PQueue) 
	    \FOR{each $G_{P} \in \mathcal{G}_P$} 
	      \IF{$G_{P}(\phi) > 0$} 
	        \STATE $k \leftarrow$ index($G_{P}$) 
	        \STATE append $N_{P}^{k}(\phi)$ to Pr
          \ENDIF
        \ENDFOR

	    \FOR{each $p\in $ Pr s.t norm(p) $> \epsilon$}
	      \FOR{each $a\in \{ N,S,E,W\}$}
	        \STATE backprop($N_{Q}^a$, $input=$ p, $target=N_{R}^a(p) + \gamma max_a(N_{Q}^a(\phi))$) 
	        \STATE Put p in PQueue with priority $|N_{Q}^a(p) - (N_{R}^a(p) + \gamma max_a(N_{Q}^a(\phi)))|$ 
            \STATE nbReplays $\leftarrow$ nbReplays + 1
          \ENDFOR
        \ENDFOR
        
    \UNTIL{PQueue empty \textbf{OR} nbReplays $\geq$ B}
  \ENDIF

\STATE $\phi^{t} \leftarrow \phi^{t+1}$ \\

  \STATE nbTrials $\leftarrow$ nbTrials $+ 1$
\UNTIL{nbTrials = maxNbTrials}
\end{algorithmic}
\end{algorithm}

\textbf{Neural Dyna-Q.} The second part of the algorithm works as a classical neural network-based Dyna-Q \cite{lin1992} with \textit{prioritized sweeping} \cite{moore1993,peng1993}. As in \cite{lin1992}, the Q-values are represented by four 2-layer feedforward neural networks $N_Q^{a}$ (one per action). During on-line phases, the agent makes decisions that drive its movements within the maze, and stores the samples in a priority queue. Their priority is the absolute value of the reward prediction error, i.e., $|\delta|$. Every time the agent receives a reward, similarly to rats, it stops and replays are simulated with a budget B (\textbf{Algo.~\ref{algo:dyna}}): the samples with the highest priority are replayed first, their potential predecessors are then estimated and placed in the queue with their respective priorities, and so on until the replay budget is exhausted.

\begin{table}[t]
  \caption{Parameter values.\label{tab:params}}
\centering
  \begin{tabular}{rc}
    \hline
    value & parameter\\
    \hline
    4000 & $maxepch$: number of epoch replays to train the world model\\
       3 & $w$: gain of the outlier detector threshold in GALMO\\
      20 & B: replay budget per stop at reward sites \\
       2 & number of layers in $N_P$, $N_R$ and $N_Q$ \\
10, 16, 26 & size of the hidden layers in $N_Q$, $N_R$ and $N_P$ (respectively)\\
$\pm0.05$, $\pm0.0045$, $\pm0.1$ & weight initialization bound in $N_Q$, $N_R$ and $N_P$ (resp.)\\
0.5, 0.1, 0.1  & learning rate in $N_Q$, $N_R$ and $N_P$ (resp.)\\
0.9, 1, 1 &  sigmoid slope in $N_P$, $N_R$ and $N_Q$ (resp.) (hidden layer)\\
0.5, 0.4, 0.4 &  sigmoid slope in $N_P$, $N_R$ and $N_Q$ (resp.) (output layer)\\

  \hline
  \end{tabular}
\end{table}

The various parameters used in the simulations are summarized in \textbf{Tab.~\ref{tab:params}}. 

\section{Results}
\subsection{Learning the world model}
\begin{figure}[t]
\centering
\includegraphics[width=.8\textwidth]{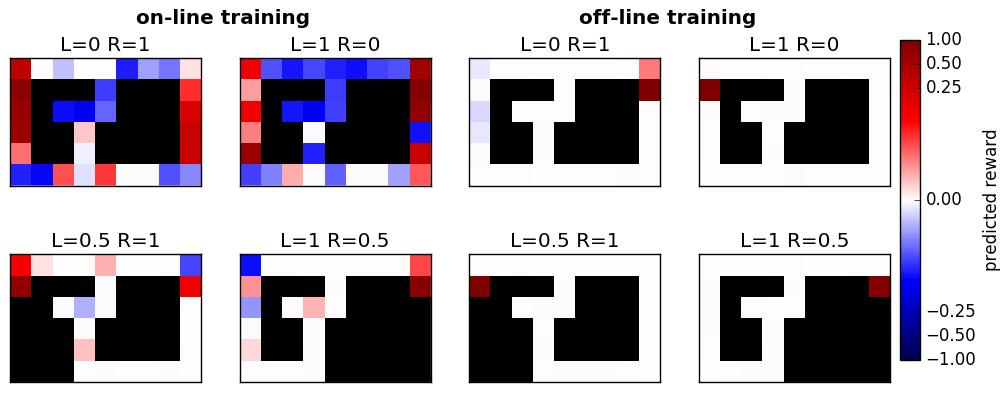}
\caption{\label{fig:rew}
\textbf{Reward predictions} are inaccurate when the model is trained on-line (Left panel) and accurate when it is trained off-line (Right panel). L, R: memory configuration. Note the use of a logarithmic scale, so as to make visible errors of small amplitude.} 
\end{figure}

Because of correlations in sample sequences, the world model is learned off-line: the samples are presented in random order, so as to break temporal correlations. We illustrate this necessity with the learning of the reward networks $N_R$: when trained on-line (\textbf{Fig.~\ref{fig:rew}, left}), the reward networks make a lot of erroneous predictions for each possible task, while when trained off-line with samples presented in randomized order, the predictions are correct (\textbf{Fig.~\ref{fig:rew}, right}).

\begin{figure}[t]
\centering
\includegraphics[width=0.9\textwidth]{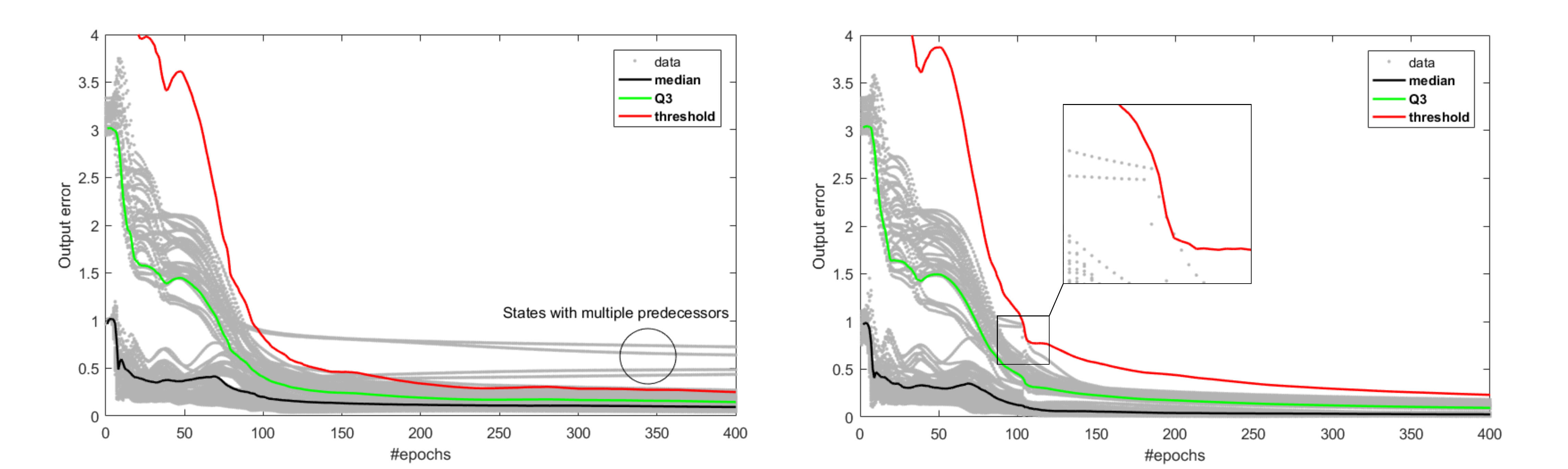}
\includegraphics[width=0.85\textwidth]{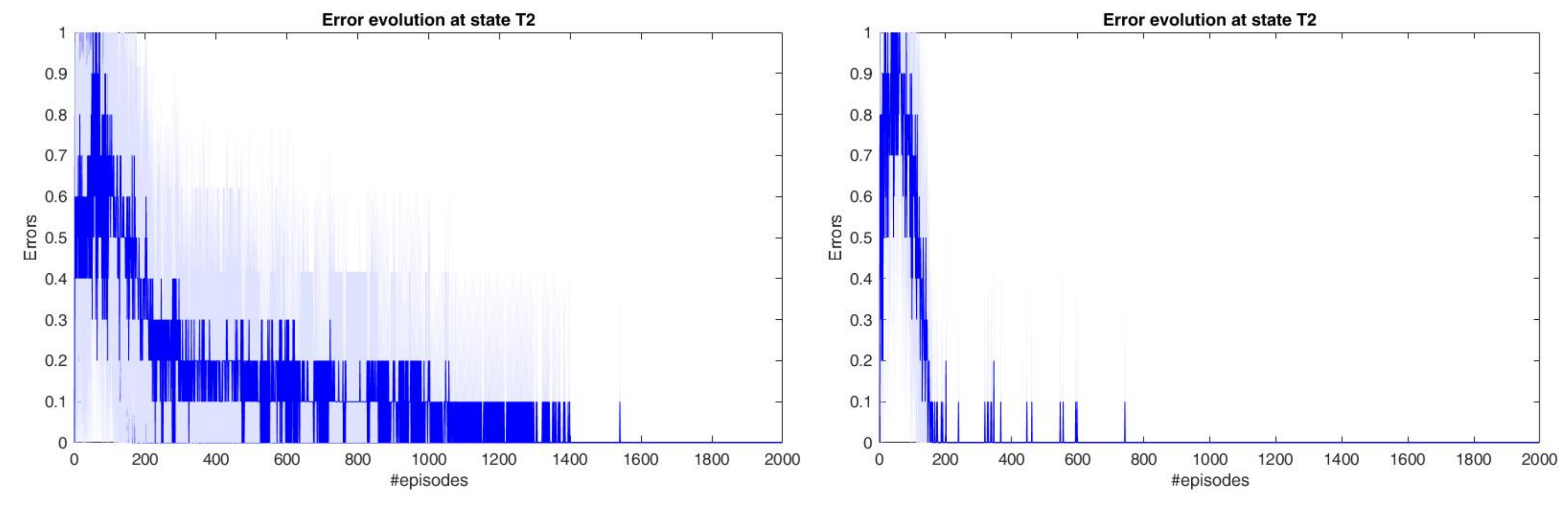}
\caption{\label{fig:GALMOtest}
\textbf{Top: Learning error dynamics without (left) and with (right) GALMO.} Errors of all samples (gray) during epochs of training. 
GALMO allows for the creation of multiple prediction networks to handle the states where multiple outputs have to be generated. \textbf{Bottom: Learning without (left) and with (right) replays.} Evolution of the proportion of decision errors at point T2 during the alternation task. Blue: 10 run mean, light blue: standard deviation.}
\end{figure}

%
With a single set of $N_P$ networks, the error of the whole set of states decreases steadily with learning, except for four states which have multiple predecessors (\textbf{Fig.~\ref{fig:GALMOtest}, top left}). With the GALMO algorithm, when the error of these states reaches the threshold $\theta$ (in red on \textbf{Fig.~\ref{fig:GALMOtest}, top right}), networks are duplicated and specialized for each of the possible predecessors. We repeated the experiment 10 times. It always converged, with a number of final networks comprised between 2 and 5 (3 times 2 networks, 1 time 3, 5 times 4, and 1 time 5). 

%

\subsection{Reinforcement learning with multiple predecessors}
%

We compare the efficiency of the Dyna-Q model we developed with the corresponding Q-learning (i.e. the same architecture without replays with the world model). As expected, Q-learning is able to learn the task, as evidenced by the proportion of erroneous choices at the decision point T2 (\textbf{Fig.~\ref{fig:GALMOtest}, bottom left})
. On average it does not fully converge before 1000 epochs of training, whereas the Dyna-Q learns much faster, thanks to the replay mechanism (\textbf{Fig.~\ref{fig:GALMOtest}, bottom right}), 
converging on average after 200 trials.

\subsection{Preliminary analysis of generated replays}

\begin{figure}
\centering
\includegraphics[width=.49\textwidth]{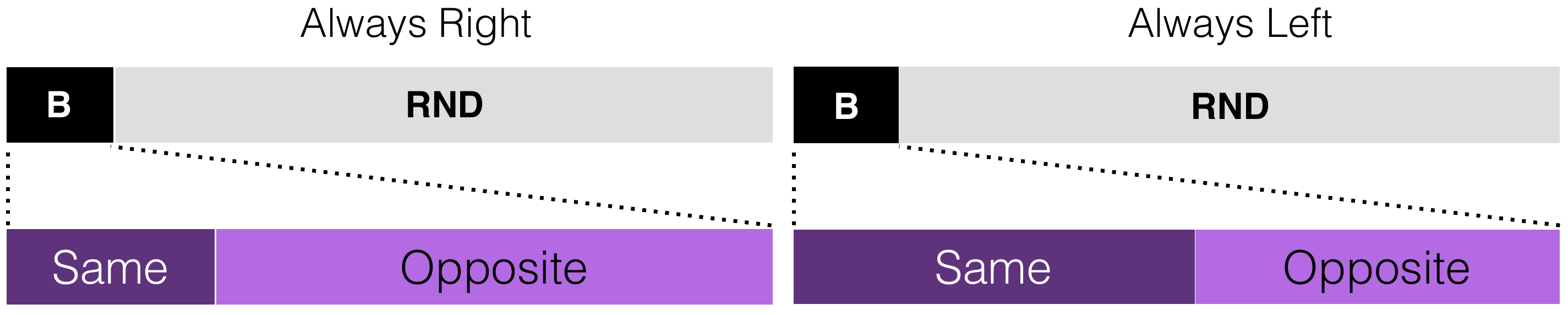}
\includegraphics[width=.49\textwidth]{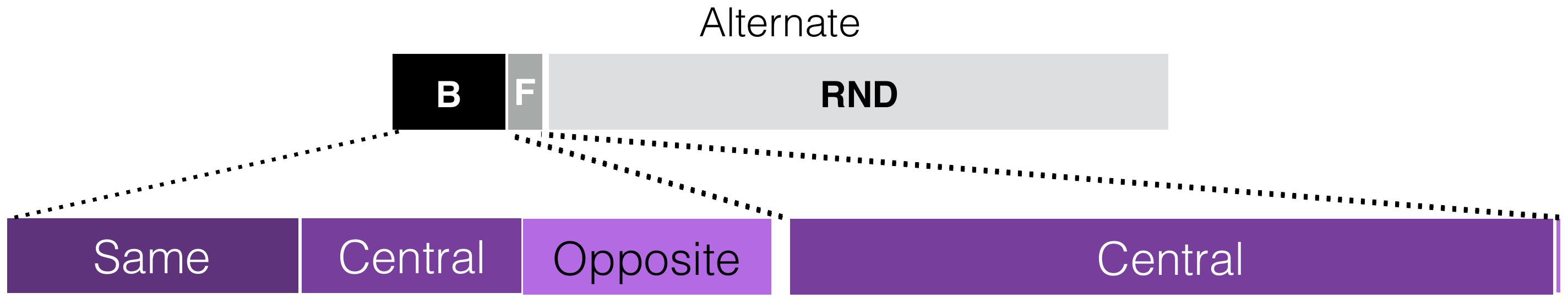}
\caption{\label{fig:perf}
\textbf{Type of replays}. B: backward, F: forward, RND: random. \label{fig:replays}}
\end{figure}

%

We analyze a posteriori the state reactivations caused by the prioritized sweeping DynaQ algorithm in always turn right, always turn left and alternate tasks. Prioritized sweeping does not rely on explicit replay of sequences, but it does favors them. 
We considered sequences or replays involving three or more consecutive steps; with 128 possible states, a 3-state sequence has a chance level of $0.01\%$ of being produced by a uniform random selection process. 
We observed in all cases (Fig.~\ref{fig:replays}) that a bit more than $80\%$ of the state reactivations did not correspond to actual sequences. Most of the sequences are backward, except for the alternate task, which also generated $4.5\%$ of forward ones. As in  \cite{Gupta2010}, we classified these sequences as being on the same side as the current agent location, on the opposite side, or in the central part of the maze. There is no clear pattern here, except that central reactivations were observed in the alternate task only.


\section{Discussion}

We proposed a new neural network architecture (GALMO) designed to associate multiple outputs to a single input, based on the multiple expert principle \cite{jacobs1991}. 
We implemented a neural version of the DynaQ algorithm \cite{lin1992}, using the prioritized sweeping principle \cite{moore1993,peng1993}, using GALMO to learn the world model. This was necessary because the evaluation task, adapted from \cite{Gupta2010}, contained some states that have multiple predecessors.

We showed that this system is able to learn the multiple predecessors cases, and to solve the task faster than the corresponding Q-learning system (i.e., without replays). This required learning the world-model off-line, with data presented in shuffled order, so as to break the sequential correlations between them, which prevented the convergence of the learning process. This result makes an interesting neuroscientific prediction (independently of the use of GALMO): if the learning principles of the rat brain are similar to those of the gradient descent for artificial neural network, then the world model has to be learned off-line, which would be compatible with non-sequential hippocampal replays. Besides, the part of the DynaQ algorithm that uses the world model to update the Q-values predicts a majority of non-sequential replays, but also $15$ to $20\%$ of sequential reactivations, both backward and forward.

Concerning GALMO, it has been tested with a quite limited set of data, and should thus be evaluated against larger sets in future work. In our specific case, the reward networks $N_R^a$ did not require the use of GALMO; a single network could learn the full $(s,a)\rightarrow r$ mapping as rewards were deterministic. But should they be stochastic, GALMO could be used to learn the multiple possible outcomes. Note that, while it has been developed in order to learn a predecessor model in a DynaQ architecture, GALMO is much more general, and would in principle be able to learn any one-to-many mapping. Finally, in the model-based and dyna reinforcement learning contexts, having multiple predecessors or successors is not an exceptional situation, especially in a robotic paradigm. The proposed approach is thus of interest beyond the task used here.



\section*{Acknowledgements}

The authors thank O. Sigaud for fruitful discussions, and F. Cinotti for proofreading. This work has received funding from the European Union’s Horizon 2020 research and innovation programme under grant agreement No 640891 (DREAM Project). This work was performed within the Labex SMART (ANR-11-LABX-65) supported by French state funds managed by the ANR within the Investissements d'Avenir programme under reference ANR-11-IDEX-0004-02.

%
%
\bibliography{biblio.bib}

\begin{thebibliography}{10}

\bibitem{Okeefe71}
O'Keefe, J., Dostrovsky, J.:
\newblock The hippocampus as a spatial map. preliminary evidence from unit
  activity in the freely-moving rat.
\newblock Brain research \textbf{34}(1) (1971)  171--175

\bibitem{Wilson1994}
Wilson, M.A., McNaughton, B.L.,  et~al.:
\newblock Reactivation of hippocampal ensemble memories during sleep.
\newblock Science \textbf{265}(5172) (1994)  676--679

\bibitem{girardeau2009}
Girardeau, G., Benchenane, K., Wiener, S.I., Buzs{\'a}ki, G., Zugaro, M.B.:
\newblock Selective suppression of hippocampal ripples impairs spatial memory.
\newblock Nature neuroscience \textbf{12}(10) (2009)  1222--1223

\bibitem{Foster2006}
Foster, D.J., Wilson, M.a.:
\newblock {Reverse replay of behavioural sequences in hippocampal place cells
  during the awake state.}
\newblock Nature \textbf{440}(7084) (2006)  680--3

\bibitem{Lee2002}
Lee, A.K., Wilson, M.A.:
\newblock {Memory of Sequential Experience in the Hippocampus during Slow Wave
  Sleep}.
\newblock Neuron \textbf{36}(6) (2002)  1183--1194

\bibitem{Gupta2010}
Gupta, A.S., van~der Meer, M.A.A., Touretzky, D.S., Redish, A.D.:
\newblock {Hippocampal Replay Is Not a Simple Function of Experience}.
\newblock Neuron \textbf{65}(5) (2010)  695--705

\bibitem{chen2017}
Chen, Z., Wilson, M.A.:
\newblock Deciphering neural codes of memory during sleep.
\newblock Trends in Neurosciences (2017)

\bibitem{peyrache2009}
Peyrache, A., Khamassi, M., Benchenane, K., Wiener, S.I., Battaglia, F.P.:
\newblock {Replay of rule-learning related neural patterns in the prefrontal
  cortex during sleep.}
\newblock Nature Neuroscience \textbf{12}(7) (2009)  919--926

\bibitem{mcclelland1995}
McClelland, J.L., McNaughton, B.L., O'reilly, R.C.:
\newblock Why there are complementary learning systems in the hippocampus and
  neocortex: insights from the successes and failures of connectionist models
  of learning and memory.
\newblock Psychological review \textbf{102}(3) (1995)  419

\bibitem{delavilleon2015}
De~Lavill{\'e}on, G., Lacroix, M.M., Rondi-Reig, L., Benchenane, K.:
\newblock Explicit memory creation during sleep demonstrates a causal role of
  place cells in navigation.
\newblock Nature neuroscience \textbf{18}(4) (2015)  493--495

\bibitem{Caze18}
Caz{\'e}, R., Khamassi, M., Aubin, L., Girard, B.:
\newblock Hippocampal replays under the scrutiny of reinforcement learning
  models.
\newblock submitted (2018)

\bibitem{sutton1990}
Sutton, R.S.:
\newblock Integrated architectures for learning, planning, and reacting based
  on approximating dynamic programming.
\newblock In: Proceedings of the seventh international conference on machine
  learning. (1990)  216--224

\bibitem{moore1993}
Moore, A.W., Atkeson, C.G.:
\newblock Prioritized sweeping: Reinforcement learning with less data and less
  time.
\newblock Machine learning \textbf{13}(1) (1993)  103--130

\bibitem{peng1993}
Peng, J., Williams, R.J.:
\newblock Efficient learning and planning within the dyna framework.
\newblock Adaptive Behavior \textbf{1}(4) (1993)  437--454

\bibitem{khamassi2005}
Khamassi, M., Lacheze, L., Girard, B., Berthoz, A., Guillot, A.:
\newblock Actor-critic models of reinforcement learning in the basal ganglia:
  from natural to arificial rats.
\newblock Adaptive Behavior \textbf{13} (2005)  131--148

\bibitem{lin1992}
Lin, L.H.:
\newblock Self-improving reactive agents based on reinforcement learning,
  planning and teaching.
\newblock Machine learning \textbf{8}(3/4) (1992)  69--97

\bibitem{paz2004}
Paz-Villagr{\'a}n, V., Save, E., Poucet, B.:
\newblock Independent coding of connected environments by place cells.
\newblock European Journal of Neuroscience \textbf{20}(5) (2004)  1379--1390

\bibitem{eichenbaum2017}
Eichenbaum, H.:
\newblock Prefrontal--hippocampal interactions in episodic memory.
\newblock Nature Reviews Neuroscience \textbf{18}(9) (2017)  547

\bibitem{Sutton1998}
Sutton, R., Barto, A.:
\newblock Reinforcement Learning: An Introduction.
\newblock Cambridge, MA: MIT Press (1998)

\bibitem{tsitsiklis1997}
Tsitsiklis, J.N., Van~Roy, B.:
\newblock Analysis of temporal-diffference learning with function
  approximation.
\newblock In: Advances in neural information processing systems. (1997)
  1075--1081

\bibitem{jacobs1991}
Jacobs, R.A., Jordan, M.I., Nowlan, S.J., Hinton, G.E.:
\newblock Adaptive mixtures of local experts.
\newblock Neural computation \textbf{3}(1) (1991)  79--87

\end{thebibliography}

\end{document}